\documentclass[a4paper]{article}

\usepackage{INTERSPEECH_v2}
\usepackage{latexsym}
\usepackage{graphicx}
\usepackage{tipa}
\usepackage{devanagari}
\usepackage{amsmath,amsthm}
\usepackage{lipsum}
\usepackage{enumerate}
\usepackage{url}
\title{A Generative Model of a Pronunciation Lexicon for Hindi\thanks{A tool for letter-to-sound conversion for Hindi under GNU license is available at an open source link "https://sourceforge.net/projects/pls-for-indic-languages/" and can be downloaded from this link for Windows and Linux machines.}}
\name{Pramod Pandey$^1$, Somnath Roy$^1$}
\address{
  $^1$Centre for Linguistics, Jawaharlal Nehru University, Delhi}
\email{pkspandey@gmail.com, somnathroy86@gmail.com}

\begin{document}

\maketitle
\begin{abstract}
Voice browser applications in Text-to- Speech (TTS) and Automatic Speech Recognition (ASR) systems crucially depend on a pronunciation lexicon. The present paper describes the model of pronunciation lexicon of Hindi developed to automatically generate the output forms of Hindi at two levels, the \textless phoneme\textgreater\space and the \textless PS\textgreater\space (PS, in short for Prosodic Structure). The latter level involves both syllable-division and stress placement. The paper describes the tool developed for generating the two-level outputs of lexica in Hindi.

\end{abstract}


\section{Introduction}

Hindi demonstrates considerable regularity in mapping the orthographic
representation to pronunciation and is thus very suitable for automatic generation of
pronunciation by a programme seeded with hand-written rules. In this paper we
describe the model developed to automatically generate output forms at two levels\textemdash
Phonemic\footnote{The term 'phonemic' is being used here in the broad sense in which it has been used in W3C
recommendations.} and Prosodic Structure (PS) \textemdash from the input forms in the Devnagari script
in which Hindi is written.\\
The paper is organized as follows. Section 2 presents the salient points of a
pronunciation lexicon relevant to the present work. Section 3 describes the
pronunciation lexicon specifications for Hindi. Section 4 briefly describes Devanagari
script and the Hindi Writing system. Section 5 gives an account of the rule-based
generative model for the generation of Hindi pronunciation lexica being discussed in
the present paper. Section 6 reports the results of the evaluation of the programmes
and section 7 presents the conclusions. 
\section{Pronunciation Lexicon}
In order for the information on pronunciation in a given language to be easily accessible for voice browser applications, the World Wide Consortium (W3C) Voice Browser Activity Lexicon Standards has recently proposed an advanced set of standards for pronunciation lexicons in the publication entitled \textit {Pronunciation Lexicon Specification (PLS) Version 1.0}.  
\subsection{Recommendations of W3C }
The specifications of the features of a pronunciation lexicon proposed in the W3C proposals are exhaustive \cite{wc2008pls}. They can be summed up as follows: \\
(1)
 \begin{itemize}
     \item  PLS is intended to be the standard format of the documents referenced by the \textless lexicon\textgreater \space element of SSML. The lexicon element is an external document that is loaded by the SSML document. There can be more than one PLS document.\\
     \item  A pronunciation lexicon is referenced by a SRGS grammar in an ASR processor. It can allow multiple pronunciations of a word in the grammar to represent different speaking styles. \\
     \item A pronunciation is specified in terms of the "alphabet" attribute of a PLS document. The main valid value of the "alphabet" attribute is the "ipa" or vendor-defined strings of the form "x-organization" or "x-organization-alphabet", as for example,"x-SAMPA". \\
     \item As a PLS processor must support "ipa" as the value of the \textless alphabet\textgreater \space attribute, it must support the Unicode representations of the phonetic characters of IPA.  These include a set of vowel and consonant symbols, a syllable delimiter, diacritics, symbols for prosodic features of stress, tone and intonation. \\
     \item A word in the PLS markup language is described by following elements. The details related to PLS markup language can be found in \cite{wc2008pls}
     \begin{itemize}
         \item The first element is \textless lexicon\textgreater. This element has many attributes namely\textemdash i. version ii. xml:base iii. xmlns iv. xml:lang v. alphabet 
        \item The second elment is \textless meta\textgreater \space or \textless metadata\textgreater. The meta element is optional and specified by attributes like name,http-equiv and content.
        \item Other necessary elements besides \textless lexicon\textgreater \space are \textless lexeme\textgreater \space, \textless grapheme\textgreater \space and \textless phoneme\textgreater. 
        \item Other optional elements are \textless alias\textgreater \space and \textless example\textgreater.
	
     \end{itemize}
     
 \end{itemize}
 
\subsection{Letter-to-Sound Rules}
The term 'Letter-to-Sound' (LTS) rules or 'Grapheme-to-Phoneme' (G2P) rules refers to rules that change orthography into pronunciation. For the suitability of the term to the Devnagari script, in which Hindi is written, a more specific term, 'Akshara-to-Sound' (ATS) rules in place of LTS or G2P rules, has been recently recommended and has come to be in use \cite{nag2014akshara,pandey2014akshara}. 

Devnagari is a prominent representative of alpha-syllabic scripts derived from Brahmi and used for major Indic languages as well as languages outside India (e.g. Tibetan, Hankul, etc.) \cite{patel2007akshara}. The linguistic level of sounds to which the Devnagari script in Hindi relates is both phonemic and phonetic.\\
The main features of A/LTS rules are discussed below.\\

For languages with irregular relations between orthographic and pronunciation levels, for instance, English, LTS rules can be generated automatically following data-driven approach. For languages with regular relations between orthographic and pronunciation levels, e.g. Hindi and Bangla, A/LTS rules can be generated by hand-written rules.\\
	
Some speech synthesis systems such as FESTIVAL \cite{black1997automatically} have provision for both data-driven and hand-written rules.\\
The basic form of a hand-written rule is the following (see http://festvox.org/bsv/x1429.html): \\
 	
(2) (LC[alpha]RC=\textgreater beta) 

Where LC is "left context on zero or more input symbols" and RC is "right context on zero or more input symbols".\\

Writing rules by hand can be very difficult, as it may require a full grasp of the factors involved in determining the pronunciations of words and longer utterances. The following points have to be taken into consideration in writing hand-written A/LTS rules.\\
(3)\\
\begin{itemize}
\item A/LTS rules may refer to word boundary but cannot refer to previous or following words.
\item Attention must be given to avoiding conflicting rules.
\item The following factors lend complexity to A/LTS rules\cite{klatt1982letter}
\begin{itemize}
\item POS 
\item Morphological boundary
\item Stress marks 
\item Exceptions due to historical change
\item Borrowings with different A/LTS rules than regular forms
\item The length of LC
\item The length of RC 
\end{itemize}
\end{itemize}

The A/LTS rules for Hindi, as we shall see below, take these factors into account and give a fairly efficient programme for converting written Hindi words into IPA representations. 

For an overview of earlier studies on Letter-to-Sound rules in general as well as Indic languages\cite{pandey2014akshara}. It should be pointed out here, however, that attempts at Letter-to-Sound rules for Hindi have been few and far between and have focused on individual aspects, but not all aspects, of the pronunciation of words. For example \cite{choudhury2002rule,narasimhan2004schwa} present comprehensive accounts of the Schwa Deletion process,\cite{hussain2004letter} presents A/LTS rules for segments in Urdu, and \cite{kishore2002data} proposes the use of the syllable for text-processing in a TTS system for Indian languages. In the model of PLS described here, all the aspects of segmental and prosodic features are included in a single program for a pronunciation lexicon of Hindi.
	
\section{Pronunciation Lexicon Specifications for Hindi}

The features included for the pronunciation lexica of Hindi are the following:\\
(4)\\
\begin{itemize}
\item Letter-to-sound conversion using IPA. 
\item Marking of syllable structures and labeling them as prominent and non-prominent for further TTS and ASR applications.
\item Preparation of the systematic variations in the pronunciations of words in Standard Hindi in two lexica, Standard Formal and Standard Colloquial varieties.
\item The surface pronunciations of words with stress marks.
\end{itemize}

The features mentioned above have been taken into account, with multiple uses in mind for the learners of Hindi, TTS and ASR applications, as well as users of screen-reading software in Hindi, among others. Two levels of pronunciations are being produced: the level of segmental IPA symbols with prosodic structure in terms of syllable division and prominence (\textless PS\textgreater), and the level of segmental IPA symbols with stress marks (\textless phoneme\textgreater), as illustrated below in the Figures 1 \& 2 in the form of screenshots of the input-output generation of two words.

\begin{figure}[h]
\centering
\includegraphics[scale=0.4]{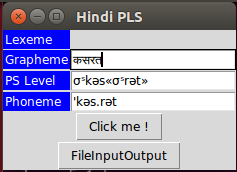}
\qquad
\includegraphics[scale=0.4]{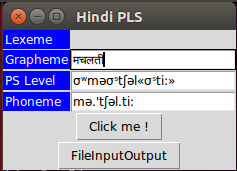}
\caption{Screenshots of the input-output generation of two words}
\end{figure}

\section{Hindi Writing and Devnagari Script}
The official script of Hindi is Devnagari or Devanagari. It belongs to a class of writing systems known as alpha-syllabic, that is, a combination of alphabetic and syllabic systems. Writing systems are alphabetic, if their units stand for phonemic/ phonetic vowels and consonants (e.g. English, German). They are syllabic, if each orthographic unit in them represents a syllable (e.g. Hankul for Korean) \cite{pandey2014akshara}. Hindi writing encapsulates in it linguistic awareness of its speakers at multiple levels- phonetic, phonological \cite{patel2007akshara,klatt1982letter,pandey2007phonology} and morphological \cite{frost2012universal,rao2012case} and lend it the character of a mix of surface and deep orthographic systems, rather than a simple surface orthographic system (see \cite{sproat2000computational}, for a discussion of the types of orthographic systems)  \\
The main features of an alpha-syllabic script are the following:\\
(5)\\
\begin{itemize}
\item The consonant letter has an inherent vowel, which is normally the mid central vowel /\textipa{@}/ or its lower counterpart /\textipa{5}/. \\
\item The inherent vowel can be deleted and the consequent form is represented with a subscript diacritic known as \textbf{halant}. The inherent vowel is assumed to give way to another vowel when added with a diacritic known as \textbf{matra} (mora). \\
\item When not preceded by a consonant, a vowel occurs alone with a full syllabic and vocalic value, e.g. {\dn aAI} /\textipa{a:i:}/ 'me-PST-FEM-SING' \\
\item When preceded by a consonant, the vowel is represented as a diacritic (a subscript, a superscript or a ‘lateral’ script along side a superscript), and is known as a \textbf{matra}, as shown in Table 1:

\begin{table}[h]
\addtolength{\tabcolsep}{-4pt}
\begin{center}
\begin{tabular}{ |p{1.65cm}|p{1.15cm}|p{1.35cm}|p{1.25cm}|p{1.6cm}|  }
 \hline
Vowel Grapheme& IPA Symbol &Unicode Value&With Onset
Consonants& Consonant shapes formed (e.g. with the
consonant
letter ‘k’)\\
 \hline
 {\dn a}   & \textipa{@}    &U+0905  & Nil & {\dn ?} +{\dn a} = {\dn k}\\
 \hline
 {\dn aA} &   \textipa{a:}  & U+0906   &{\dn A} & {\dn ?} +{\dn aA} = {\dn kA}\\
 \hline
 {\dn i} &\textipa{i} & U+0907 &  { \dn E} &  {\dn ?} +{\dn i} = {\dn Ek}\\
 \hline
 {\dn I}    &\textipa{i:} &U+0908 & { \dn F} &  {\dn ?} +{\dn I} = {\dn kF}\\
 \hline
\end{tabular}
\end{center}
\caption{\label{Examples of four Hindi vowels in isolation and following onsets} Examples of four Hindi vowels in isolation and following onsets}
\end{table}
\item Two or three consonant clusters form ligatures, with one of the consonants (usually the last, except when it is /r/) occurring full but the others occurring half (e.g. {\dn (s} /ts/, {\dn -=l} /spl/) or occasionally forming a new grapheme, e.g. {\dn /} (from {\dn (} + {\dn r}) /t+r/.\\
\item In addition, the following superscript "\space{ {\dn \1}}\space" called chandrabindu is used for nasal vowels. Another superscript "\space{\space{\dn \2}}\space", a bindu (translated as ‘point’) has two values- (a) a homorganic nasal consonant, and (b) a nasal vowel. The use of the bindu for both nasal consonants as well as nasal vowels has lent indeterminacy in the orthography phonology conversion. The orthography requires lexical knowledge of its use. 
\end{itemize}

\section{Essential background to the Model}

\subsection{Variety of Hindi}
There are two related lexica for two styles of Standard Hindi being described here- Standard Formal Hindi (SFH) and Standard Colloquial Hindi (SCH). They are, however, in a subset relation. SCH is a superset of SFH. SCH contains mainly two points of difference in segmental realization of words from SFH, as stated in (6) below.\\
(6)\\
\begin{itemize}
\item SFH maintains vowel length distinction between /i i:/ and /u u:/ throughout in the word; SCH neutralizes it word-finally, e.g. 
\begin{table}[h]
\addtolength{\tabcolsep}{-3pt}
\begin{center}
\begin{tabular}{ p{2cm} p{2cm} p{3cm}  }
 \hline
SFH & SCH &Gloss\\
 \hline
 \textipa{@"tit\super hi} &\textipa{" @tit\super hi:} &guest\\
 \hline
  \textipa{s@"ma:d\super hi} &\textipa{s@"ma:d\super hi:} &trance/ mausoleum\\
 \hline
\end{tabular}
\end{center}
\end{table}\\
\item Word-internal [\textipa{@}] is in general deleted under certain circumstances in SCH (discussed at length in \cite{ohala1983aspects,narasimhan2004schwa,choudhury2002rule}, but is retained in SFH, e.g. 
\begin{table}[h]
\addtolength{\tabcolsep}{-3pt}
\begin{center}
\begin{tabular}{ p{2cm} p{2cm} p{3cm}  }
 \hline
SFH & SCH &Gloss\\
 \hline
 \textipa{"k@m@la:} &\textipa{" k@mla:} &'a name'\\
 \hline
  \textipa{"kit@ni:} &\textipa{'kitni:} &'how much-fem'\\
 \hline
\end{tabular}
\end{center}
\end{table}
\end{itemize}

\subsection{Rule-based generation of outputs}

The central focus of the present programme was the generation rather then listing of
the entries of lexemes in the lexicon. In order to achieve this end a well-ordered set of
rules was formulated in three separate steps.\\
The first set of rules was of correspondence between graphemes and their IPA
equivalents. All the graphemic shapes were taken into account- as single consonant
and vowel letters, and as CV, CCV and CCCV aksharas, e.g., {\dn a} /\textipa{@}/, {\dn kF} /\textipa{ki:}/, {\dn ?yA} /\textipa{kja:}/,{\dn -/F} /\textipa{stri:}/\\
The second set of rules was regarding the changes in the grapheme- phoneme
correspondence on account of segmental processes. These included 
(a) final vowel lengthening, whereby all short vowel graphemes had to correspond to their long
vowel counterparts, e.g. {\dn aEtET} /\textipa{@tit\super hi}/  $\rightarrow$ /\textipa{@tit\super hi:}/, and (b) word-final schwa deletion , e.g. {\dn kml} /\textipa{k@m@l@}/ $\rightarrow$ \textbf{/\textipa{k@m@l}/}
The third set of rules consisted of prosodic structure rules. These were mainly rules
that involved division of words into syllables and labeling of syllables. The labeling
was in two stages, in turn. In the first stage, labeling was based on three degrees of
weight \cite{kelkar1968studies,pandey1990hindi,hayes1995metrical}, namely Light (CV), Heavy (CVV/CVC) and Superheavy
(CVCC/ CVVC), which were labeled as $\sigma$\super w, $\sigma$\super s  and   $\sigma$\super{s'}, respectively.The second stage,
the relabeling stage, was consequent on the following phenomena \textemdash (a) word-internal
schwa deletion \cite{pandey1989word}, which targeted /\textipa{@}/ in $\sigma$\super w syllables, e.g. /\textipa{k@.m@.la:}/ $\rightarrow$  /\textipa {k@m.la:}/, (b) Demotion or downgrading of $\sigma$\super s syllables into $\sigma$\super w syllables on account of stress clash, e.g. /$\sigma$\super sa:$\sigma$\super{s'}sa:n/ $\rightarrow$ /$\sigma$\super wa: $\sigma$\super{s'}sa:n/ 'easy',(c) and promotion or upgrading of$\sigma$\super w syllables
as $\sigma$\super s in disyllabic words with the first syllables as Light syllables, as in /\textipa{"k@la:}/. There
is one more metrical feature that was incorporated in the prosodic structure of words,
namely, Extrametricality \cite{hayes1995metrical} of final heavy syllables. The final extrametrical \textless $\sigma$\super s\textgreater \space syllables were ignored for stress. All other
$\sigma$\super s and $\sigma$\super{s’} were converted into stress at the \textless phoneme\textgreater\space level. The significance of
Extrametricality of the final heavy syllable is that in the speech synthesis of longer
utterances, the word-final heavy syllable can be optionally de-extrametricalized and
stressed, e.g. [\textipa{"kit@ni:}] \textgreater [\textipa{"kitni:}] 'how much- FEM' (focused).\\

An earlier programme of PLS took into account the phenomenon of consonant
gemination as part of the second set of rules. Consonants in Hindi are geminated or lengthened when preceding /r, l, \textipa{V}, j/. The rule can be stated as (C+r/l/\textipa{V}/j $\rightarrow$ C1C+r/l/\textipa{V}/j), e.g., /\textipa{m\ae tri:}/$\rightarrow$ [\textipa{m\ae ttri:}] 'friendship' and /\textipa{a:V@Sj@k}/ $\rightarrow$ and [\textipa{a:V@SSj@k}] 'necessary'. This feature is underspecified in the orthogrpahy \cite{pandey2007phonology}. It has been
left out of the final version of the model on the grounds of its suitability to TTS and
ASR systems. We find that the duration of geminates /tt ll/, etc. is not the double of
singlets, but of less duration. For example, it was found that the duration of singlet /t/
and /l/ was 0.96 and 0.91 respectively, and of geminates of 1.10 and 1.06 ms. in
doublets such as /\textipa{p@ta:}/ ‘address’ and /\textipa{p@tta:}/ ‘leaf’, and /\textipa{b@la:}/ ‘trouble’ /\textipa{b@lla:}/
‘bat(N)’. In such a case the duration of the geminate consonants can be roughly
predicted by rule for the given contexts. Besides, the programme was simpler without
the rule of gemination.\\

The full derivations of four forms {\dn ksrt} /\textipa{k@s@r@t@}/ 'exercise' and {\dn mcltF}
/\textipa{m@\:tS@l@ti:}/ 'prancing', {\dn p\5\9kEt} /\textipa{pr@kriti}/ 'nature' and {\dn sAlAnA} /\textipa{sa:la:na:} 'yearly' by applying the A/LTS rules of Hindi are given in Table-2
below. ‘Gr’ stands for ‘Grapheme’.

\begin{table*}[h]
\addtolength{\tabcolsep}{-6pt}
\begin{center}
\begin{tabular}{ |p{2cm}|p{3.5cm}|p{3.8cm}|p{3.8cm}|p{3.8cm}|  }
 \hline
Rule Applied& Gr-1:{\dn ksrt} Output-1& Gr-2:{\dn mcltF} Output-2 & Gr-3:{\dn p\5\9kEt} Output-3 & Gr-4: {\dn sAlAnA} \space output-4 \\
 \hline
ATS Rules-1,2,3 Akshara-IPA Correspondence rules & /\textipa{k@s@r@t@}/ & /\textipa{m@\:tS@l@ti:}/ & /\textipa{pr@kriti}/ & /\textipa{sa:la:na:}/\\
 \hline
 ATS Rule-4 Final Schwa Deletion   &[\textipa{k@s@r@t}] & DNA & DNA & DNA\\
 \hline
 
 ATS Rule-5 Final Vowel Lengthening   &DNA & DNA& [\textipa{pr@kriti:}]  & DNA\\
 \hline
  ATS Rule-6 Syllabification &  [\textipa{$\sigma$k@$\sigma$s@$\sigma$r@t}] & [\textipa{$\sigma$m@$\sigma$\:tS@$\sigma$l@$\sigma$ti:}] &[\textipa{$\sigma$pr@$\sigma$ kri$\sigma$ti:}] & [\textipa{$\sigma$sa:$\sigma$la:$\sigma$na:}]\\
 \hline
 ATS Rule-7 Syllable Labeling by Weight & [\textipa{$\sigma$\super wk@$\sigma$\super ws@ $\sigma$\super sr@t}] &  [\textipa{$\sigma$\super wm@$\sigma$\super w\:tS@$\sigma$\super wl@$\sigma$\super sti:}] &[\textipa{$\sigma$\super w pr@ $\sigma$\super w kri  $\sigma$\super s ti:} & [\textipa{$\sigma$\super s sa: $\sigma$\super s la:  $\sigma$\super s na:}] \\
 \hline
  ATS Rule-8 Syllable Extrametricality & [\textipa{$\sigma$\super wk@$\sigma$\super ws@\textless$\sigma$\super sr@t\textgreater}] &  [\textipa{$\sigma$\super wm@$\sigma$\super w\:tS@$\sigma$\super wl@$\sigma$\super sti:}] &[\textipa{$\sigma$\super w pr@ $\sigma$\super w kri \textless $\sigma$\super s ti:\textgreater} & [\textipa{$\sigma$\super s sa: $\sigma$\super s la: \textless $\sigma$\super s na:\textgreater}] \\
 \hline
  ATS Rule-9 Syllable Relabeling, Downgrading & DNA & DNA & DNA & [\textipa{$\sigma$\super w sa: $\sigma$\super s la: \textless $\sigma$\super s na:\textgreater}] \\
 \hline
 ATS Rule-10: Syllable Relabeling, Upgrading  & [\textipa{$\sigma$\super sk@$\sigma$\super ws@\textless$\sigma$\super sr@t\textgreater}] &  [\textipa{$\sigma$\super wm@$\sigma$\super s\:tS@$\sigma$\super wl@$\sigma$\super sti:}] &[\textipa{$\sigma$\super s pr@ $\sigma$\super w kri \textless $\sigma$\super s ti:\textgreater}] & DNA\\
 \hline
 ATS Rule-11:  Internal Schwa Deletion & [\textipa{$\sigma$\super sk@$\sigma$\super ws\textless$\sigma$\super sr@t\textgreater}] &  [\textipa{$\sigma$\super wm@$\sigma$\super s\:tS@$\sigma$\super wl$\sigma$\super sti:}] &DNA & DNA\\
 \hline
  ATS Rule-12: Resyllabification &[\textipa{$\sigma$\super sk@s\textless$\sigma$\super sr@t\textgreater}] &  [\textipa{$\sigma$\super wm@$\sigma$\super s\:tS@l$\sigma$\super sti:}] &DNA & DNA\\
 \hline
 ATS Rule 14-15: Leveling and Stress Mark Assignment & [\textipa{"k@sr@t}] &  [\textipa{m@"\:tS@lti:}] & [\textipa{" pr@kriti:}] & [\textipa{sa:"la:na:}] \\
 \hline
\end{tabular}
\end{center}
\caption{\label{Rule Application and output} Rule Application and output}
\end{table*}

 \subsection{Treatment of exceptions}
 
In order for efficient automatic generation of orthographic forms of Hindi words into
prosodic structure and IPA symbols by means of hand-written rules in the model, it
was necessary to include an Exceptions component which is fed with the output of
given words that are not in conformity with the rest of the outputs (see \cite{ramakrishnan2007grapheme} with a
similar suggestion for Tamil). These are the following in the main:\\

(7)\\

a. \textbf{Native vocabulary}

i. Limited sets of forms with some prefixes and compounds without
internal word-boundaries. For example, /\textipa{su-s@m@j}/ 'good time' and
/\textipa{ku-s@m@j}/ 'bad time', in which the schwa in the first syllable will be wrongly deleted- *[\textipa{"susm@j}] and *[\textipa{"kusm@j}] in place of [\textipa{"sus@m@j}] and [\textipa{"kus@m@j}].

ii. Non-honorific Imperative forms of verbs ending in the suffix /-a:/. For
example, /\textipa{di"kha:}/ 'show-CAUS+IMP+NH' versus /\textipa{"dikha:}/ 'was seen'.

iii. Complex words with O-O correspondence relation with simple words.
For example, the plural forms of feminine nouns /\textipa{m@\:tS\super h@li:}/ 'fish' and /\textipa{tit@li:}/ 'butterfly' as /\textipa{m@\:tS\super h@lij\~a:}/ 'fish-PL' and /\textipa{tit@lij\~a:}/ 'butterfly-PL'. The output forms with stress following the regular A/LTS rules would
be *[\textipa{m@\:"tS\super h@lij\~a:}] *[\textipa{ti"t@lij\~a:}]. The preferred forms are [\textipa{"m@\:tS\super hlij\~a:}] and [\textipa{"titlij\~a:}] instead. The latter are placed in the Exceptions component.

iv. The use of the diacritics of bindu and chandrabindu in Hindi
orthography for two phenomena- vowel nasalization and nasal
consonant. Although they are in complementary distribution in general,
with the chandrabindu standing for a nasal vowel, and the bindu
standing for a nasal consonant, as in {\dn P\1sA}/\textipa{p\super h\~@sa:}/ 'trapped (PASS)' {\dn P\2dA} /\textipa{ph@nda:}/ 'noose', in actual use there is a lot of free variation.

b. \textbf{Borrowed English vocabulary}: words borrowed from English containing short vowels but non-distinctive diacritic (matra) for the vowels [e e:], e.g. {\dn \u kAl\?j} 'college'. The rule-based form here would be *[\textipa{"kOle:dZ}]. The regular pronounced form is [\textipa{"kOledZ}].

\section{Evaluation}

The PLS programme for Standard Colloquial Hindi was taken for evaluation, as it
contained more features for evaluation than the programme for Standard Formal Hindi, as discussed in section 5.1.
The data used for evaluation came from a BBC word corpus containing 28731
words. An expert for the pronunciations of the words first annotated the words and
then the output of the machine was compared against the annotated list. The
following factors were taken into account in evaluating the model:\\
a. Only the regular features of pronunciation were considered.\\
b. A transcription was considered correct if it contained one of the main features of
akshara-to-sound correspondence in the expected context, as described. \\
c. In evaluating the model for the regular features of pronunciation, the factors
relating to exceptions had to be excluded.\\
Out of the total number of 28731 words tested, the number of words with the
diacritics bindu and chandrabindu for the realization of nasal consonants and nasal
vowels was 3705 and with deletable schwas 14460. For stress marking as well as for
the akashara-to-phoneme correspondence, the entire corpus of 28731 words was
considered relevant. For the marking of stress, although only polysyllabic words were
relevant, we had to ensure that monosyllabic words were not marked redundantly for
stress. Thus the entire corpus of 28731 words was relevant for stress. The result of the
analysis of errors is presented in Table 3 below:

\begin{table}[h]
\addtolength{\tabcolsep}{-6pt}
\begin{center}
\begin{tabular}{ |l|l|l|l| }
\hline
 {The Phenomena Tested} & {\# Words} & {\# Wrong Outputs} & {Percentage Accuracy} \\\hline
 {Nasal} & {3705} & {82} & {97.79} \\\hline
 {PS Structure} & {28731} & {254} & {99.12} \\\hline
 {Schwa Deletion} & {14460} & {116} & {99.20} \\\hline
 {Anusvara/Anunasika} &{700} & {2} & {99.72} \\\hline
 {ATS Correspondence} & {28731} & {0} & {100} \\\hline
\end{tabular}
\end{center}
\caption{\label{Performance of the proposed system} Performance of the proposed system}
\end{table}

\section{Conclusion}
Developing a programme for generating a pronunciation lexicon for standard Hindi
involved coming to terms with several issues described above. The programme began
with generating initial grapheme to phoneme correspondences. However, in an
attempt to automatically generate the final output at the \textless phoneme\textgreater\space level it was
found that the mixed nature of the Hindi orthography with both surface and deep
correspondences made it necessary to automatically generate the internal word-
prosodic structure on which the main aspects of segmental realization of forms
depended.
The present model of pronunciation lexicon for Hindi can be easily applied to other
akshara-based orthographic systems with mixed level orthographies. In addition, the
contributions extend beyond the scope of a pronunciation lexicon and have general
applicability for automating syllable-based processes and stress patterns in South
Asian languages.

\bibliographystyle{IEEEtran}

\bibliography{mybib}


\end{document}